\documentclass[conference]{IEEEtran}

\usepackage{graphicx} 
\usepackage{subfigure} 

\usepackage[numbers,sort&compress]{natbib}

\usepackage{algorithm}
\usepackage{algorithmic}

\usepackage{amsmath}
\usepackage{amssymb}
\usepackage{amsthm}
\usepackage{comment}

\newenvironment{remark}[1][Remark]{\begin{trivlist}
\item[\hskip \labelsep {\bfseries #1}]}{\end{trivlist}}

\providecommand{\ie}{\emph{i.e.,} }
\providecommand{\eg}{\emph{e.g.,} }
\DeclareMathOperator*{\minimize}{minimize}

\newtheorem{lemma}{Lemma}

\providecommand{\ie}{\emph{i.e.,} }
\providecommand{\eg}{\emph{e.g.,} }

\providecommand{\mypara}[1]{\smallskip\noindent\emph{#1} }
\providecommand{\myparab}[1]{\smallskip\noindent\textbf{#1} }

\begin{document} 
\title{
Stock Market Prediction from WSJ:\\Text Mining via Sparse Matrix Factorization}

\author{
\IEEEauthorblockN{Felix Ming Fai Wong, Zhenming Liu, Mung Chiang}
\IEEEauthorblockA{Princeton University\\
mwthree@princeton.edu, zhenming@cs.princeton.edu, chiangm@princeton.edu
}
}

\maketitle

\begin{abstract} 
We revisit the problem of predicting directional movements of stock prices based on news articles: here our algorithm uses daily articles from The Wall Street Journal to predict the closing stock prices on the same day. We propose a unified latent space model to characterize the ``co-movements'' between stock prices and news articles. Unlike many existing approaches, our new model is able to simultaneously leverage the correlations: (a) among stock prices, (b) among news articles, and (c) between stock prices and news articles. Thus, our model is able to make daily predictions on more than 500 stocks (most of which are not even mentioned in any news article) while having low complexity. We carry out extensive backtesting on trading strategies based on our algorithm. The result shows that our model has substantially better accuracy rate (55.7\%) compared to many widely used algorithms. The return (56\%) and Sharpe ratio due to a trading strategy based on our model are also much higher than baseline indices. 
\end{abstract}

\section{Introduction}
A main goal in algorithmic trading in financial markets is to 
predict if a stock's price will go up or down at the end of the current trading day as the  algorithms continuously receive new market information. 
One variant of the question is to construct effective prediction algorithms based on news articles. Understanding this question is important for two reasons: (1) A better solution 
helps us gain more insights on how financial markets react to news, which 
is a long-lasting question in finance~\cite{fama98,chan03,tetlock07}. (2) It presents 
a unique challenge in machine learning, where  time series analysis meets text information retrieval. 
While there have been quite extensive studies on stock price prediction based on news,
much less work can be found on \emph{simultaneously} leveraging the correlations (1) among stock prices, (2) among news articles, and (3) between stock prices and news articles~\cite{minev12}. 

In this paper, we revisit the stock price prediction problem based on news articles.
On each trading day, we feed a prediction algorithm all the articles that appeared on that day's Wall Street Journal (WSJ) (which becomes available before the market opens), then we ask the algorithm  to predict whether each stock in S\&P 500, DJIA and Nasdaq will move up or down.
Our algorithm's accuracy is approximately 55\% (based on $\geq 100,000$ test cases). This shall be contrasted with ``textbook models'' for time series that have less than $51.5\%$ prediction accuracy (see Section~\ref{sec:evaluation}). We also remark that we require the algorithm to predict \emph{all} the stocks of interest while most of the stocks are \emph{not mentioned at all} in a typical WSJ newspaper. On the other hand, most of the existing news-based prediction algorithms can predict only stocks that are explicitly mentioned in the news. Finally, when we use this algorithm to construct a portfolio, we find our portfolio yields substantially better return and Sharpe ratio compared to a number of standard indices
(see Figure \ref{fig:returns_1213}). 

\myparab{Performance surprises.} We were quite surprised by the performance of our algorithm for the following reasons. 

\mypara{(1) Our algorithm runs on minimal data.} Here, we use only daily open and close prices and WSJ news articles.
It is clear that
all serious traders on Wall Street have access to both pieces of information, and much more.
By the efficient market hypothesis, it should be difficult 
to find arbitrage based on our dataset (in fact, the efficient market hypothesis explains why the accuracy rates of ``textbook models'' are
 below $51.5\%$). Thus, we were intrigued by the performance of our algorithm. It also appears that the market might not 
be as efficient as one would imagine. 

\mypara{(2) Our model is quite natural but it appears to have never been studied before.} As we shall see in the forthcoming sections, 
our model is rather natural for capturing the correlation between stock price movements and news articles.
While the news-based stock price prediction problem has been extensively studied~\cite{minev12},
we have not seen a model similar to ours in existing literature.
Section~\ref{sec:related} also compares our model with a number of important existing approaches. 

\mypara{(3) Our algorithm is robust.}  Many articles in WSJ are on events that happened a day before (instead of reporting new stories developed overnight).  Intuitively, the market shall 
be able to absorb information immediately and thus ``old news'' should be excluded from a prediction algorithm. Our algorithm does not attempt to filter out any news since deciding the freshness of a news article appears to be remarkably difficult, and yet even when a large portion 
of the input is not news, our algorithm can still make profitable predictions. 

\myparab{Our approach.} 
We now outline our solution. We build a unified latent factor model to explain stock price movements and news. Our model originates from straightforward ideas in time series analysis and information retrieval: when we study co-movements of multiple stock prices, we notice that the price movements can be embedded into a low dimensional space. The low dimensional space can be ``extracted'' using standard techniques such as Singular Value Decomposition. On the other hand, when we analyze texts in news articles, it is also standard to embed each article into latent spaces using techniques such as 
probabilistic latent semantic analysis or latent Dirichlet allocation \cite{murphy12}.

Our crucial observation here is that stock prices and financial news should ``share'' the same latent space.  
For example, the coordinates of the space can represent stocks' and news articles' weights on different industry sectors
(\eg technology, energy) and/or topics (\eg social, political).
Then if a fresh news article is about ``crude oil," we should see a larger fluctuation in the prices of stocks with higher
weight in the ``energy sector" direction.
 
Thus, our approach results in a much simpler and more interpretable model. But even in this simplified model, 
we face a severe overfitting problem: we use daily trading data over six years. 
Thus, there are only in total approximately 1500 trading days. 
On the other hand, we need to predict about 500 stocks. When the dimension of our latent space is only ten,
we already have 5000 parameters. In this setting, appropriate regularization is needed. 

Finally, our inference problem involves non-convex optimization. We use Alternating Direction Method of Multipliers (ADMM) \cite{boyd11} to solve the problem. Here the variables in the ADMM solution are matrices and thus we need a more general version of ADMM. While the generalized analysis is quite straightforward, it does not seem to have appeared in the literature. This analysis for generalized ADMM could be of independent interest. 

In summary,
\begin{enumerate}
\item We propose a unified and natural model to leverage the correlation between stock price movements and news articles. This model allows us to predict the prices of all the stocks of interest even when 
most of them are not mentioned in the news.  
\item We design appropriate regularization mechanisms to address the overfitting problem and develop a generalized ADMM algorithm for inference. 
\item We carry out extensive backtesting experiments to validate the efficacy of our algorithm. We also compare our algorithm with a number of widely used models and observe substantially improved performance. 
\end{enumerate}

\section{Notation and Preliminaries}

Let there be $n$ stocks, $m$ words, and $s+1$ days (indexed as $t=0,1,\ldots,s$). We then define the following variables:
\begin{itemize}
\item $x_{it}$: closing price of stock $i$ on day $t$,
\item $y_{jt}$: intensity of word $j$ on day $t$,
\item $\displaystyle r_{it} = \log\Bigl(\frac{x_{it}}{x_{i,t-1}}\Bigr)$: log return of stock $i$ on day $t\geq 1$.
\end{itemize}

The stock market prediction problem using newspaper text is formulated as follows:
\textbf{
for given day $t$, use both historical data $[r_{it'}]$, $[y_{jt'}]$ (for $t' \leq t$) and this morning's newspaper $[y_{jt}]$ to predict $[r_{it}]$, for all $i$ and $j$.}\footnote{$[x_{it}]$ is recoverable from $[r_{it}]$ given $[x_{i,t-1}]$ is known.}

In this paper we compute $y_{jt}$  as the z-score on the number of newspaper articles that contain word $j$
relative to the article counts in previous days. To reduce noise, an extra thresholding step is included to remove values
that are negative or below 3 standard deviations.

\myparab{Dataset.}
We use stock data in a period of almost six years and newspaper text from WSJ.

We identified 553 stocks that were traded from 1/1/2008 to 9/30/2013 and listed in at 
least one of the S\&P 500, DJIA, or Nasdaq stock indices during that period.
We then downloaded opening and closing prices\footnote{We adjust prices for stock splits, but do not account for dividends in our evaluation.} of the stocks from CRSP.\footnote{CRSP, Center for Research in Security Prices. Graduate School of Business, The University of Chicago 2014. Used with permission. All rights reserved. www.crsp.uchicago.edu}
Additional stock information was downloaded from Compustat.
For text data, we downloaded the full text of all articles published in the print version of WSJ in the same period.
We computed the document counts per day that mention the top 1000 words of highest frequency and the company names of the 553 stocks. 
After applying a stoplist and removing company names with too few mentions, we obtained a list of 1354 words.

\section{Sparse Matrix Factorization Model}
Equipped by recent advances in matrix factorization techniques for collaborative filtering~\cite{koren09},
we propose a unified framework that incorporates
(1) historical stock prices, (2) correlation among different stocks and (3) newspaper content
to predict stock price movement.
Underlying our technique is a latent factor model that characterizes a stock
(\eg it is an energy stock)
and the \emph{average} investor mood of a day (\eg economic growth in America becomes more robust and thus the demand for energy is projected to increase),
and that the price of a stock on a certain day is a function of the latent features of 
the stock and the investor mood of that day.

More specifically, we let stocks and trading days share a $d$-dimensional latent factor space, so that
stock $i$ is described by a nonnegative feature vector $u_i \in \mathbb{R}_+^d$
and trading day $t$ is described by another feature vector $v_t \in \mathbb{R}^d$.
Now if we assume $u_i$ and $v_t$ are known, we model day $t$'s log return, $\hat{r}_{it}$, as the
inner product of the feature vectors $\hat{r}_{it} = u_i^T v_t + \epsilon$, where $\epsilon$ is a noise term. 
In the current setting we can only infer $v_t$ by that morning's newspaper articles as described by
$y_t = [y_{jt}] \in \mathbb{R}_+^m$, so naturally we may assume a linear transformation $W\in \mathbb{R}^{d\times m}$
to map $y_t$ to $v_t$, \ie we have $v_t = W y_t$. Then log return prediction can be expressed as
\begin{equation}
\hat{r}_{it} = u_i^T W y_t.
\end{equation}

Our goal is to learn the feature vectors $u_i$ and mapping $W$ using historical data from $s$ days.
Writing in matrix form: let
$R = [r_{it}] \in \mathbb{R}^{n\times s}$, $U = [u_1 \cdots u_n]^T \in \mathbb{R}^{n\times d}$, $Y = [y_1 \cdots y_s] \in \mathbb{R}^{m\times s}$,
we aim to solve
{
\begin{equation}
\minimize_{U\geq 0,\ W}\qquad \frac{1}{2} \lVert R - UWY \rVert_{F}^2.
\end{equation} 
}
\vspace{-.2cm}
\begin{remark}Here, the rows of $U$ are the latent variables for the stocks while the columns of 
$WY$ are latent variables for the news. We allow one of $U$ and $WY$ to be negative to reflect the fact that
news can carry negative sentiment while we force the other one  to be non-negative to control the complexity of the model. Also, the model becomes less interpretable when both $U$ and $WY$ can be negative. 
\end{remark}

Note our formulation is similar to the standard matrix factorization problem except we add the matrix $Y$.
Once we have solved for $U$ and $W$ we can predict price $\hat{x}_{it}$ for day $t$ by
$\hat{x}_{it} = x_{i,t-1} \exp(\hat{r}_{it}) = x_{i,t-1} \exp\bigl( u_i^T W y_t\bigr)$ 
given previous day's price $x_{i,t-1}$ and the corresponding morning's newspaper word vector $y_t$.

\myparab{Overfitting.}
We now address the overfitting problem. Here, we introduce the following two additonal requirements to our model:
\begin{enumerate}
\item We require the model to be able to produce a predicted log returns matrix $\hat{R} = [\hat{r}_{it}]$ that is close to $R$ and be of low rank at the same time, and
\item be sparse because we expect many words to be irrelevant to stock market prediction (a feature selection problem) and
each selected word to be associated with few factors.
\end{enumerate}

The first requirement is satisfied if we set $d\ll s$.
The second requirement motivates us to  introduce a sparse group lasso \cite{friedman10} regularization term
in our optimization formulation. More specifically, feature selection means we want only a small number of columns of $W$ (each column corresponds to one word) to be nonzero, and this can be induced by introducing the regularization term
$\lambda \sum_{j=1}^m \lVert W_j \rVert_2$,
where $W_j$ denotes the $j$-th column of $W$ and $\lambda$ is a regularization parameter.
On the other hand, each word being associated with few factors means that
for each relevant word, we want its columns to be sparse itself. This can be induced by introducing the regularization term
$\mu \sum_{j=1}^n \lVert W_j \rVert_1 = \mu \lVert W \rVert_1$,
where $\mu$ is another regularization parameter, and $\lVert W \rVert_1$ is taken elementwise.

Thus our optimization problem becomes
\begin{align}
\minimize_{U,\ W}&\quad \frac{1}{2} \lVert R - UWY \rVert_{F}^2 + \lambda \sum_{j=1}^{m} \lVert W_j \rVert_2 + \mu \lVert W \rVert_1 \nonumber\\
\text{subject to} &\quad U\geq 0. \label{opt_prob}
\end{align}

We remark we also have examined other regularization approaches, \eg $\ell_2$ regularization and plain group lasso, but they do not outperform
baseline algorithms. Because of space constraints, this paper focuses on understanding the performance of the current approach.

\section{Optimization Algorithm}
Our problem is biconvex, \ie convex in either $U$ or $W$ but not jointly. 
It has been observed such problems can be effectively solved by ADMM~\cite{zhang10_admm}. 
Here, we study how such techniques can be applied in our setting.
We rewrite the optimization problem by replacing the nonnegative constraint with an indicator function and
introducing auxiliary variables $A$ and $B$:
\begin{align}
\minimize_{A,\ B,\ U,\ W}\quad &\frac{1}{2} \lVert R - ABY \rVert_{F}^2 + \lambda \sum_{j=1}^{m} \lVert W_j \rVert_2 \nonumber \nonumber \\
& + \mu \lVert W \rVert_1 + I_{+}(U) \nonumber \\
\text{subject to} \quad &A = U,\ B = W,  
\end{align}
where $I_{+}(U) = 0$ if $U\geq 0$, and $I_{+}(U) = \infty$ otherwise.

We introduce Lagrange multipliers $C$ and $D$ and formulate the augmented Lagrangian of the problem:
\begin{align}
&L_{\rho}(A,B,U,W,C,D) \nonumber \\
&=  \frac{1}{2} \lVert R - ABY \rVert_F^2 + \lambda \sum_{j=1}^m \lVert W_j \rVert_2
+ \mu \lVert W \rVert_1 + I_{+}(U) \nonumber \\
& \quad + \text{tr}\bigl(C^T (A-U)\bigr) + \text{tr}\bigl(D^T (B-W)\bigr) \nonumber \\
& \quad + \frac{\rho}{2} \lVert A-U \rVert_F^2 + \frac{\rho}{2} \lVert B-W \rVert_F^2.
\end{align}

Using ADMM, we iteratively update the variables $A$, $B$, $U$, $W$, $C$, $D$,
such that in each iteration (denote $G_+$ as the updated value of some variable $G$):
\begin{align*}
A_+ &= \text{argmin}_{A} \ L_{\rho}(A,B,U,W,C,D) \nonumber\\
B_+ &= \text{argmin}_{B} \ L_{\rho}(A_+,B,U,W,C,D) \nonumber\\
U_+ &= \text{argmin}_{U} \ L_{\rho}(A_+,B_+,U,W,C,D) \nonumber\\
W_+ &= \text{argmin}_{W} \ L_{\rho}(A_+,B_+,U_+,W,C,D)\\
C_+ &= C + \rho (A_+ - U_+)\\
D_+ &= D + \rho (B_+ - W_+).
\end{align*}

Algorithm \ref{pseudocode} lists the steps involved in ADMM optimization, and the remaining of this section presents the detailed derivation of the update steps.

\begin{algorithm}
\caption{ADMM optimization for \eqref{opt_prob}.}
\label{pseudocode}
\renewcommand{\algorithmicrequire}{\textbf{Input:}}
\renewcommand{\algorithmicensure}{\textbf{Output:}}
\begin{algorithmic}
\REQUIRE $R,Y,\lambda,\mu,\rho$
\ENSURE $U,W$
\STATE Initialize $A,B,C,D$
\REPEAT
	\STATE $A \leftarrow (RY^TB^T - C + \rho U) (BYY^TB^T + \rho I)^{-1}$
	\STATE $B \leftarrow \text{solution to } $
	\STATE \hspace{0.5cm} $ \Bigl(\frac{1}{\rho} A^TA\Bigr) B (YY^T) + B = \frac{1}{\rho} (A^TRY^T - D) + W$
	\STATE $U \leftarrow \Bigl(A + \frac{1}{\rho}C\Bigr)^+$
	\FOR{$j=1$ to $m$}
		\STATE $\displaystyle W_j \leftarrow \biggl(\frac{\lVert w \rVert_2 - \lambda}{\rho \lVert w \rVert_2}\biggr)^+ w$, where 
		\STATE \hspace{1cm} $w = \rho\ \text{sgn}(v) (|v| - \mu/\rho)^+$, $v = B_j + {D_j}/{\rho}$
	\ENDFOR
	\STATE $C \leftarrow C + \rho (A - U)$
	\STATE $D \leftarrow D + \rho (B - W)$
\UNTIL{convergence or max iterations reached}
\end{algorithmic}
\end{algorithm}

Making use of the fact $\lVert G \rVert_F^2 = \text{tr}(G^TG)$,
we express the augmented Lagrangian in terms of matrix traces:
\begin{align*}
L_{\rho} &= \frac{1}{2} \text{tr}\bigl((R-ABY)^T (R-ABY)\bigr) + \lambda \sum_{j=1}^m \lVert W_j \rVert_2 
 + \mu \lVert W \rVert_1 \\ 
&\quad + I_+(U) + \text{tr}\bigl(C^T(A-U)\bigr) + \text{tr}\bigl(D^T(B-W)\bigr) \\
&\quad+ \frac{\rho}{2} \text{tr}\bigl((A-U)^T(A-U)\bigr) + \frac{\rho}{2} \text{tr}\bigl((B-W)^T(B-W)\bigr),
\end{align*}
then we expand and take derivatives as follows.

\myparab{Updating $A$.} We have
\begin{align*}
\frac{\partial L_{\rho}}{\partial A} &= 
\frac{1}{2}\frac{\partial \text{tr}(Y^TB^TA^TABY)}{\partial A} - \frac{1}{2} \cdot 2 \frac{\partial \text{tr}(R^TABY)}{\partial A} \\
&\quad + \frac{\partial \text{tr}(C^TA)}{\partial A} + \frac{\rho}{2} \frac{\partial \text{tr}(A^TA)}{\partial A} - \frac{\rho}{2} \cdot 2 \frac{\partial \text{tr}(U^TA)}{\partial A} \\
&= ABYY^TB^T - RY^TB^T + C + \rho A - \rho U.
\end{align*}
By setting the derivative to $0$, the optimal $A^*$ satisfies
\begin{align*}
A^* &= (RY^TB^T - C + \rho U) (BYY^TB^T + \rho I)^{-1}.
\end{align*}

\myparab{Updating $B$.} Similarly,
\begin{align*}
\frac{\partial L_{\rho}}{\partial B} &=
\frac{1}{2}\frac{\partial \text{tr}(Y^TB^TA^TABY)}{\partial B} - \frac{1}{2} \cdot 2 \frac{\partial \text{tr}(R^TABY)}{\partial B} \\
&\quad + \frac{\partial \text{tr}(D^TB)}{\partial B} + \frac{\rho}{2} \frac{\partial \text{tr}(B^TB)}{\partial B} - \frac{\rho}{2} \cdot 2 \frac{\partial \text{tr}(W^TB)}{\partial B},
\end{align*}
then setting $0$ and rearranging, we have
\begin{align*}
\Bigl(\frac{1}{\rho} A^TA\Bigr) B^* (YY^T) + B^* &= \frac{1}{\rho} (A^TRY^T - D) + W.
\end{align*}
Hence $B^*$ can be computed by solving the above Sylvester matrix equation of the form $AXB + X = C$.

\myparab{Solving matrix equation $AXB + X = C$.} 
To solve for $X$, we apply the Hessenberg-Schur method \cite{golub79} as follows:
\begin{enumerate}
\item Compute $H = U^TAU$, where $U^TU=I$ and $H$ is upper Hessenberg, \ie $H_{ij} = 0$ for all $i>j+1$.
\item Compute $S = V^TBV$, where $V^TV=I$ and $S$ is quasi-upper triangular, \ie $S$ is triangular except with possible $2 \times 2$ blocks along the diagonal.
\item Compute $F = U^TCV$.
\item Solve for $Y$ in $HYS^T + Y = F$ by back substitution.
\item Solve for $X$ by computing $X=UYV^T$.
\end{enumerate}

To avoid repeating the computationally expensive Schur decomposition step (step 2), we precompute and store the
results for use across multiple iterations of ADMM. This prevents us from using a one-line call to numerical packages
(\eg $\mathtt{dlyap()}$ in Matlab) to solve the equation.

Here we detail the back substitution step (step 4), which was omitted in \cite{golub79}.
Following \cite{golub79}, we use $m_k$ and $m_{ij}$ to denote the $k$-th column and $(i,j)$-th element of matrix $M$ respectively.
 Since $S$ is quasi-upper triangular, we can solve for $Y$ from the last column,
and then back substitute to solve for the second last column, and so on.
The only complication is when a $2\times 2$ nonzero block exists; in that case we
solve for two columns simultaneously. More specifically:

(a) If $s_{k,k-1} = 0$, we have
\begin{align*}
H\biggl(\sum_{j=k}^n s_{kj} y_j\biggr) + y_k &= f_k\\
(s_{kk} H + I) y_k &= f_k - H\sum_{j=k+1}^n s_{kj} y_j,
\end{align*}
then we can solve for $y_k$ by Gaussian elimination.

(b) If $s_{k,k-1} \ne 0$, we have
\begin{align*}
H \begin{bmatrix} y_{k-1} & y_k \end{bmatrix} \begin{bmatrix}s_{k-1,k-1} & s_{k,k-1} \\ s_{k-1,k} & s_{kk} \end{bmatrix} + \begin{bmatrix} y_{k-1} & y_k \end{bmatrix} & \\
= \begin{bmatrix} f_{k-1} & f_k \end{bmatrix} - \sum_{j=k+1}^n H \begin{bmatrix} s_{k-1,j}y_j & s_{kj}y_j \end{bmatrix} &.
\end{align*}

The left hand side can be rewritten as
\begin{align*}
H \begin{bmatrix} s_{k-1,k-1}y_{k-1} + s_{k-1,k} y_k & s_{k,k-1}y_{k-1} + s_{kk}y_k \end{bmatrix} + \\ \begin{bmatrix} y_{k-1} & y_k \end{bmatrix}\\
= [ (s_{k-1,k-1}H+I)y_{k-1} + s_{k-1,k}Hy_k \cdots  \\  s_{k,k-1}Hy_{k-1} + (s_{kk}H+I)y_k ] \\
= \begin{bmatrix} s_{k-1,k-1}H+I & s_{k-1,k}H \\ s_{k,k-1}H & s_{kk}H+I \end{bmatrix}\begin{bmatrix} y_{k-1} \\ y_k \end{bmatrix}
\end{align*}
by writing $\begin{bmatrix} y_{k-1} & y_k \end{bmatrix}$ as $\begin{bmatrix} y_{k-1} \\ y_k \end{bmatrix}$. 
The right hand side can also be rewritten as
\begin{align*}
\begin{bmatrix} f_{k-1} \\ f_k \end{bmatrix} - \sum_{j=k+1}^n \begin{bmatrix} s_{k-1,j} H y_j \\ s_{kj} H y_j \end{bmatrix}.
\end{align*}

Thus we can solve for columns $y_k$ and $y_{k-1}$ at the same time through Gaussian elimination on
\begin{align*}
\begin{bmatrix} s_{k-1,k-1}H+I & s_{k-1,k}H \\ s_{k,k-1}H & s_{kk}H+I \end{bmatrix} 
\begin{bmatrix} y_{k-1} \\ y_k \end{bmatrix} & \\
 = \begin{bmatrix} f_{k-1} \\ f_k \end{bmatrix} - \sum_{j=k+1}^n \begin{bmatrix} s_{k-1,j} H y_j \\ s_{kj} H y_j \end{bmatrix} &.
\end{align*}

\myparab{Updating $U$.} Note that
\begin{align*}
U_+ &= \text{argmin}_U \ I_+(U) - \text{tr}(C^TU) + \frac{\rho}{2} \lVert A- U\rVert_F^2 \\
&= \text{argmin}_U \ I_+(U) + \frac{\rho}{2} \biggl\lVert \Bigl(A+\frac{1}{\rho}C\Bigr) - U \biggr\rVert_F^2 \\
&= \Bigl(A + \frac{1}{\rho}C\Bigr)^+,
\end{align*} 
with the minimization in step 2 being equivalent to taking the Euclidean projection onto the convex set of nonnegative matrices  \cite{boyd11}.

\myparab{Updating $W$.} $W$ is chosen to minimize
\begin{align*}
\lambda \sum_{j=1}^m \lVert W_j \rVert_2 + \mu \lVert W \rVert_1 - \text{tr}(D^T W) + \frac{\rho}{2} \lVert B-W \rVert_F^2. 
\end{align*}
Note that this optimization problem can be solved for each of the $m$ columns of $W$ separately:
\begin{align}
W_j^* &= \text{argmin}_u \ \lambda \lVert u \rVert_2 + \mu \lVert u \rVert_1 - D_j^T u + \frac{\rho}{2} \lVert B_j - u \rVert_2^2 \nonumber\\
&= \text{argmin}_u \ \lambda \lVert u \rVert_2 + \mu \lVert u \rVert_1 + \frac{\rho}{2} \biggl\lVert u - \Bigl(B_j + \frac{D_j}{\rho}\Bigr) \biggr\rVert_2^2, \label{w_min}
\end{align}
We can obtain a closed-form solution by studying the subdifferential of the above expression.
\begin{lemma}
\label{lemma_min}
Let $F(u) = \lambda \lVert u \rVert_2 + \mu \lVert u \rVert_1 + \rho/2 \lVert u-v \rVert_2^2$. Then the minimizer $u^*$ of $F(u)$ is
\begin{align*}
u^* = \biggl(\frac{\lVert w \rVert_2 - \lambda}{ \rho \lVert w \rVert_2}\biggr)^+ w,
\end{align*}
where $w = [w_i] $ is defined as $w_i = \rho \ \text{sgn}(v_i) (|v_i| - \mu/\rho)^+$.
\end{lemma}

This result was given in a slightly different form in \cite{sprechmann11}. We include a more detailed proof here for completeness.

\begin{proof}
$u^*$ is a minimizer iff $0\in \partial F(u^*)$, where
\begin{align*}
\partial F(u) &= \lambda \partial \lVert u \rVert_2 + \mu \partial \lVert u \rVert_1 + \nabla \frac{\rho}{2} \lVert u-v \rVert_2^2, \text{ with } \\
\partial \lVert u \rVert_2 &= \begin{cases} \displaystyle \Bigl\{\frac{u}{\lVert u \rVert_2}\Bigr\} & u \ne 0 \\ \{s \mid \lVert s \rVert_2 \leq 1\} & u = 0\end{cases}\\
\partial \lVert u \rVert_1 &= [\partial |u_i|]\\
\partial |u_i| &= \begin{cases} \{\text{sgn}(u_i)\} & u_i \ne 0 \\ [-1,1] & u_i = 0. \end{cases}
\end{align*}

In the following, $\lVert \cdot \rVert$ denotes $\lVert \cdot \rVert_2$, and $\text{sgn}(\cdot)$, $|\cdot|$, $(\cdot)^+$ are understood to be done elementwise if operated on a vector.
There are two cases to consider:

Case 1: $\lVert w \rVert \leq \lambda$

This implies $u^* = 0$, $\partial \lVert u^* \rVert_2 = \{s \mid \lVert s \rVert \leq 1\}$, $\partial \lVert u^* \rVert_1 = \{t \mid t\in [-1,1]^n\}$, and $\nabla \lVert u^* -v\rVert_2^2 = - \rho v$. Then
\begin{align*}
0 \in \partial F(u^*) &\iff  0 \in \{\lambda s + \mu t - \rho v \mid \lVert s \rVert \leq 1, t \in [-1,1]^n\}\\
&\iff \exists s:\ \lVert s \rVert \leq 1,\ t \in [-1,1]^n \\
&\qquad \text{ s.t. } \biggl(\lambda s + \mu t = \rho v \iff v - \frac{\mu}{\rho} t = \frac{\lambda}{\rho} s\biggr).
\end{align*}
Now we show an $(s,t)$ pair satisfying the above indeed exists.
Define $t = [t_i]$ such that 
\begin{align*}
t_i = \begin{cases}\displaystyle \frac{\rho}{\mu} v_i & \displaystyle |v_i| \leq \frac{\mu}{\rho}, \\ \text{sgn}(v_i) & \displaystyle |v_i|>\frac{\mu}{\rho}. \end{cases}
\end{align*}
If $|v_i| \leq \mu/\rho$, then $\rho/\mu(-\mu/\rho) \leq t_i \leq \rho/\mu(\mu/\rho) \Rightarrow t_i \in [-1,1]$. If $|v_i| > \mu/\rho$, then obviously $t_i \in [-1,1]$. Therefore we have $t \in [-1,1]^n$.

Now define $s = (\rho v - \mu t)/\lambda$. We first write
\begin{align*}
\rho\ \text{sgn}(v_i) |v_i| - \mu t_i &= \begin{cases} \displaystyle \rho v_i - \mu \Bigl(\frac{\rho}{\mu} v_i\Bigr) & \displaystyle |v_i| \leq \frac{\mu}{\rho} \\ \rho\ \text{sgn}(v_i) |v_i| - \mu\ \text{sgn}(v_i) & \displaystyle |v_i| > \frac{\mu}{\rho} \end{cases} \\
&= \begin{cases} 0 & \displaystyle |v_i| \leq \frac{\mu}{\rho} \\ \displaystyle \rho\ \text{sgn}(v_i) \Bigl(|v_i| - \frac{\mu}{\rho} \Bigr) & \displaystyle |v_i| > \frac{\mu}{\rho}\end{cases}\\
&= \rho\ \text{sgn}(v_i) \Bigl(|v_i| - \frac{\mu}{\rho}\Bigr)^+.
\end{align*}
Then we show $\lVert s \rVert \leq 1$:
\begin{align*}
\lVert s \rVert &= \frac{1}{\lambda} \lVert \rho v - \mu t \rVert\\
&= \frac{1}{\lambda} \lVert \rho\ \text{sgn}(v) |v| - \mu t \rVert\\
&= \frac{1}{\lambda} \biggl\lVert \rho\ \text{sgn}(v) \Bigl( |v| - \frac{\mu}{\rho}\Bigr)^+ \biggr\rVert\\
&= \frac{1}{\lambda} \lVert w \rVert
  \leq 1.
\end{align*}
Hence we have shown $0 \in \partial F(u^*)$ for $\lVert w \rVert \leq \lambda$.

Case 2: $\lVert w \rVert > \lambda$

Here $\lVert w \rVert - \lambda > 0$ and we have $u^* = (\lVert w \rVert - \lambda)/(\rho \lVert w \rVert)\cdot w$.
Since $\lVert w \rVert \ne 0$ means $w \ne 0$, we also have $u^* \ne 0$.

Then $\partial \lVert u^* \rVert_2 = \{u/\lVert u \rVert\}$ and
\begin{align*}
\partial F(u^*) &= \Bigl\{ \frac{\lambda}{\lVert u^* \rVert} u^* + \rho (u^*- v) \Bigr\} + \mu \partial \lVert u^* \rVert_1 \\
&= \biggl\{ \biggl( \frac{\rho \lambda}{\lVert w \rVert - \lambda} + \rho\biggr) u^* - \rho v \biggr\} + \mu \partial \lVert u^* \rVert_1,
\end{align*} 
where the last step makes use of $\lVert u^* \rVert = (\lVert w \rVert - \lambda)/(\rho \lVert w \rVert) \cdot \lVert w \rVert = (\lVert w \rVert - \lambda)/\rho$.

Our goal is to show $0 \in \partial F(u^*)$, which is true iff it is valid elementwise, \ie
\begin{align*}
0 \in \partial F_i(u^*) &= \biggl\{ \biggl( \frac{\rho \lambda}{\lVert w \rVert - \lambda} + \rho\biggr) u_i^* - \rho v_i \biggr\} + \mu \partial |u_i^*|.
\end{align*}
We consider two subcases of each element $u_i^*$.

(a) The case $u_i^* = 0$ results from $w_i = 0$, which in turn results from $|v_i| \leq \mu/\rho$. Then
\begin{align*}
\partial F_i(u^*) &= \biggl\{ \biggl( \frac{\rho \lambda}{\lVert w \rVert - \lambda} + \rho\biggr)\cdot 0 - \rho v_i \biggr\} + \mu \partial |0| \\
&= \{ \mu s - \rho v_i \mid s \in [-1,1]\}\\
&= [-\mu -\rho v_i, \mu - \rho v_i].
\end{align*}
Note that for all $v_i$ with $|v_i|\leq \mu/\rho$ the above interval includes $0$, since
\begin{align*}
-\mu - \rho v_i \leq -\mu - \rho \Bigl(-\frac{\mu}{\rho}\Bigr) = 0 &\\
\mu - \rho v_i \geq \mu - \rho \Bigl(\frac{\mu}{\rho}\Bigr) = 0 &.
\end{align*}
Thus $0 \in \partial F_i(u^*)$.

(b) The case $u_i^* \ne 0$ corresponds to $|v_i| > \mu/\rho$. Then
\begin{align*}
&\partial F_i(u^*) \\ &= \biggl\{ \biggl( \frac{\rho \lambda}{\lVert w \rVert - \lambda} + \rho\biggr)u_i^* - \rho v_i \biggr\} + \{ \mu\ \text{sgn}(u_i^*) \} \\
&= \biggl\{ \frac{\rho \lVert w \rVert}{\lVert w \rVert - \lambda} u_i^* - \rho v_i + \mu\ \text{sgn}(v_i) \biggr\} \\ 
&= \biggl\{ \frac{\rho \lVert w \rVert}{\lVert w \rVert - \lambda} \frac{\lVert w \rVert - \lambda}{\rho \lVert w \rVert} \rho\ \text{sgn}(v_i) \Bigl(|v_i| - \frac{\mu}{\rho} \Bigr) 
- \rho v_i + \mu\ \text{sgn}(v_i) \biggr\}\\
&= \{ \rho v_i - \mu \text{sgn}(v_i) - \rho v_i + \mu\ \text{sgn}(v_i)\} = \{ 0 \},
\end{align*}
where the second step comes from $\text{sgn}(u_i^*) = \text{sgn}(v_i)$  by definition of $u_i^*$.
Hence $0 \in \partial F_i(u^*)$ for $\lVert w \rVert > \lambda$.
\end{proof}

Applying Lemma \ref{lemma_min} to \eqref{w_min}, we obtain
\begin{align*}
W_j^* = \biggl(\frac{\lVert w \rVert_2 - \lambda}{\rho \lVert w \rVert_2}\biggr)^+ w,
\end{align*}
where $\displaystyle w = \rho\ \text{sgn}(v) \Bigl(|v| - \frac{\mu}{\rho}\Bigr)^+$ and $\displaystyle v = B_j + \frac{D_j}{\rho}$.

\section{Evaluation}\label{sec:evaluation}

We split our dataset into a training set using years 2008 to 2011 (1008 trading days), a validation set using 2012 (250 trading days),
and a test set using the first three quarters of 2013 (188 trading days).
In the following, we report on the results of both 2012 (validation set) and 2013 (test set),
because a comparison between the two years reveals interesting insights.
We fix $d=10$, \ie ten latent factors, in our evaluation.

\subsection{Price Direction Prediction}
First we focus on the task of using one morning's newspaper text to predict the closing price of a stock on the same day.
Because our ultimate goal is to devise a profitable stock trading strategy,
our performance metric is the accuracy in predicting the up/down direction of price movement,
averaged across all stocks and all days in the evaluation period.

We compare our method with baseline models outlined below. The first two baselines are trivial models but in practice it is observed that they yield small least square prediction errors. 
\begin{itemize}
\item \textbf{Previous $X$}: we assume stock prices are flat, \ie we always predict today's closing prices being the same as yesterday's closing prices. 
\item \textbf{Previous $R$}: we assume returns $R$ are flat, \ie today's returns are the same as the previous day's returns. Note we can
readily convert between predicted prices $\hat{X}$ and predicted returns $\hat{R}$.
\item \textbf{Autoregressive (AR) models} on historical prices (``AR on $X$") and returns (``AR on $R$"): we varied the order of the AR models
and found them to give best performance at order 10, \ie a prediction depends on previous ten day's prices/returns.
\item \textbf{Regress on $X$/$R$}: we also regress on previous day's prices/returns on \emph{all} stocks to predict a stock's price/return 
to capture the correlation between different stocks.
\end{itemize}

\begin{table}
\small
\begin{center}
\caption{\small Results of price prediction.}
\label{table:predict}
\begin{tabular}{c|c c}
\hline
Model & Accuracy '12 (\%) & Accuracy '13 (\%)\\
\hline
Ours & \textbf{53.9} & \textbf{55.7} \\
Previous $X$ & 49.9 & 46.9 \\
Previous $R$ & 49.9 & 49.1 \\
AR(10) on $X$ & 50.4 & 49.5 \\
AR(10) on $R$ & 50.6 & 50.9 \\
Regress on $X$ & 50.2 & 51.4 \\
Regress on $R$ & 48.9 & 50.8 \\
\hline
\end{tabular}
\end{center}
\vspace{-0.5cm}
\end{table}

Table~\ref{table:predict} summarizes our evaluation results in this section.
Our method performs better than all baselines in terms of directional accuracy.
Although the improvements look modest by only a few percent, we will see in the next section that they result in significant financial gains.
Note that our accuracy results should not be directly compared to other results in existing work
because the evaluation environments are different. Factors that affect evaluation results include
timespan of evaluation (years vs weeks), size of data (WSJ vs multiple sources),
frequency of prediction (daily vs intraday) and
target to predict (all stocks in a fixed set vs news-covered stocks or stock indices).

\myparab{Stocks not mentioned in WSJ.}
The performance of our algorithm does not degrade over stocks that are \emph{rarely mentioned} in WSJ: Figure 1 presents a scatter
plot on stocks' directional accuracy against their number of mentions in WSJ. One
can see that positive correlations between accuracy and frequencies of mention
do not exist. To our knowledge, none of the existing prediction algorithms have
this property.

\begin{figure}
\centering
\includegraphics[width=0.45\textwidth]{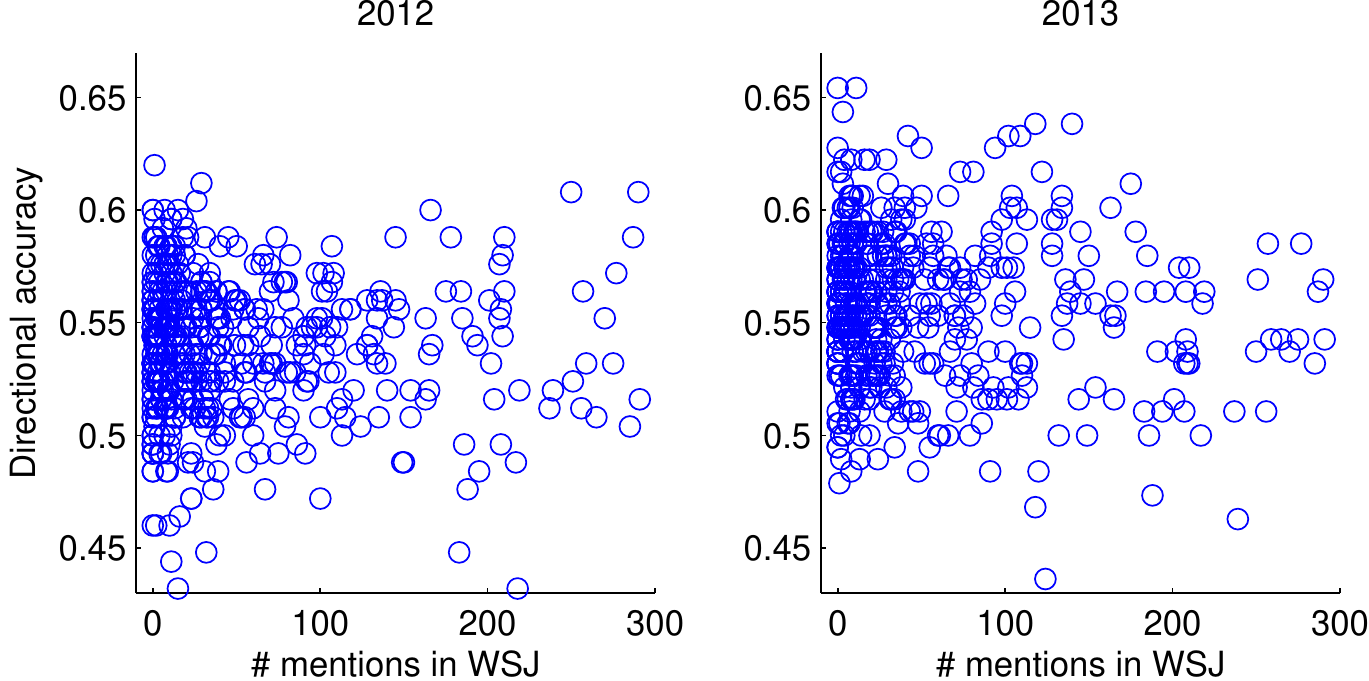}
\caption{\small Scatter plot of directional accuracy per stock.}
\label{fig:stock_accrate}
\end{figure}

\subsection{Backtesting of Trading Strategies}
We next evaluate trading strategies based on our prediction algorithm. 
We consider the following simplistic trading strategy: at the morning of each day we predict the closing prices of all stocks,
and use our current capital to buy all stocks with an ``up" prediction, such that all bought stocks have the same amount of investment.
Stocks are bought \emph{at the opening prices} of the day.
At the end of the day we sell all we have to obtain the capital for the next morning.\footnote{Incorporating shorting and transaction costs is future work.}

We compare our method with three sets of baselines: 
\begin{itemize}
\item Three major stock indices (S\&P 500, DJIA and Nasdaq), 
\item Uniform portfolios, \ie spend an equal amount of capital on each stock, and 
\item Minimum variance portfolios (MVPs) \cite{markowitz52} with expected returns at 95th percentile 
of historical stock returns.
\end{itemize}

For the latter two we consider the strategies of buy and hold (BAH), \ie buy stocks on the first day of the evaluation period
and sell them only on the last day, and constant rebalancing (CBAL), \ie for a given portfolio (weighting) of stocks we maintain the stock weights by selling and rebuying on each day.
Following \cite{ganesh13} (and see the discussion therein for the choices of the metrics), we use five performance metrics:
cumulative return, worst day return $= \min_t (X_{it} - X_{i,t-1})/X_{i,t-1}$, maximum drawdown, Conditional Value at Risk (CVaR) at 5\% level, and daily Sharpe ratio with S\&P 500 returns as reference.

Tables \ref{table:trading_2012} and \ref{table:trading_2013} summarizes our evaluation.
In both years our strategy generates significantly higher returns than all baselines.
As for the other performance metrics, our strategy dominates all baselines in 2013, and in 2012, our strategy's metrics are either the best or close to the best results.

\begin{table*}
\small
\begin{center}
\caption{\small Results of simulated trading in 2012.}
\label{table:trading_2012}
\begin{tabular}{c|c c c c c}
\hline
Model & Return & Worst day & Max drawdown & CVaR & Sharpe ratio\\
\hline
Ours & \textbf{1.21} & -0.0291 & \textbf{0.0606} & \textbf{-0.0126} & 0.0313 \\
S\&P 500 & 1.13 & -0.0246 & 0.0993 & -0.0171 & -- \\
DJIA & 1.07 & \textbf{-0.0236} & 0.0887 & -0.0159 & -0.109 \\
Nasdaq & 1.16 & -0.0282 & 0.120 & -0.0197 & \textbf{0.0320} \\
U-BAH & 1.13 & -0.0307 & 0.134 & -0.0204 & 0.00290 \\
U-CBAL & 1.13 & -0.0278 & 0.0869 & -0.0178 & -0.00360 \\
MVP-BAH & 1.06 & -0.0607 & 0.148 & -0.0227 & -0.0322 \\
MVP-CBAL & 1.09 & -0.0275 & 0.115 & -0.0172 & -0.0182 \\
\hline
\end{tabular}
\end{center}
\end{table*}

\begin{table*}
\small
\begin{center}
\caption{\small Results of simulated trading in 2013.}
\label{table:trading_2013}
\begin{tabular}{c | c c c c c}
\hline
Model & Return & Worst day & Max drawdown & CVaR & Sharpe ratio\\
\hline
Ours & \textbf{1.56} & \textbf{-0.0170} & \textbf{0.0243} & \textbf{-0.0108} & \textbf{0.148} \\
S\&P 500 & 1.18 & -0.0250 & 0.0576 & -0.0170 & -- \\
DJIA & 1.15 & -0.0234 & 0.0563 & -0.0151 & -0.0561 \\
Nasdaq & 1.25 & -0.0238 & 0.0518 & -0.0179 & 0.117 \\
U-BAH & 1.22 & -0.0296 & 0.0647 & -0.0196 & 0.0784 \\
U-CBAL & 1.14 & -0.0254 & 0.0480 & -0.0169 & -0.0453 \\
MVP-BAH & 1.24 & -0.0329 & 0.0691 & -0.0207 & 0.0447 \\
MVP-CBAL & 1.10 & -0.0193 & 0.0683 & -0.0154 & -0.0531 \\
\hline
\end{tabular}
\end{center}
\end{table*}

\section{Interpretation of the Models and Results.}
\begin{figure}
\centering
\subfigure[t-SNE on rows of $U$. Each stock is a datapoint and each color represents an NAICS industry sector.]{
\includegraphics[width=0.2\textwidth]{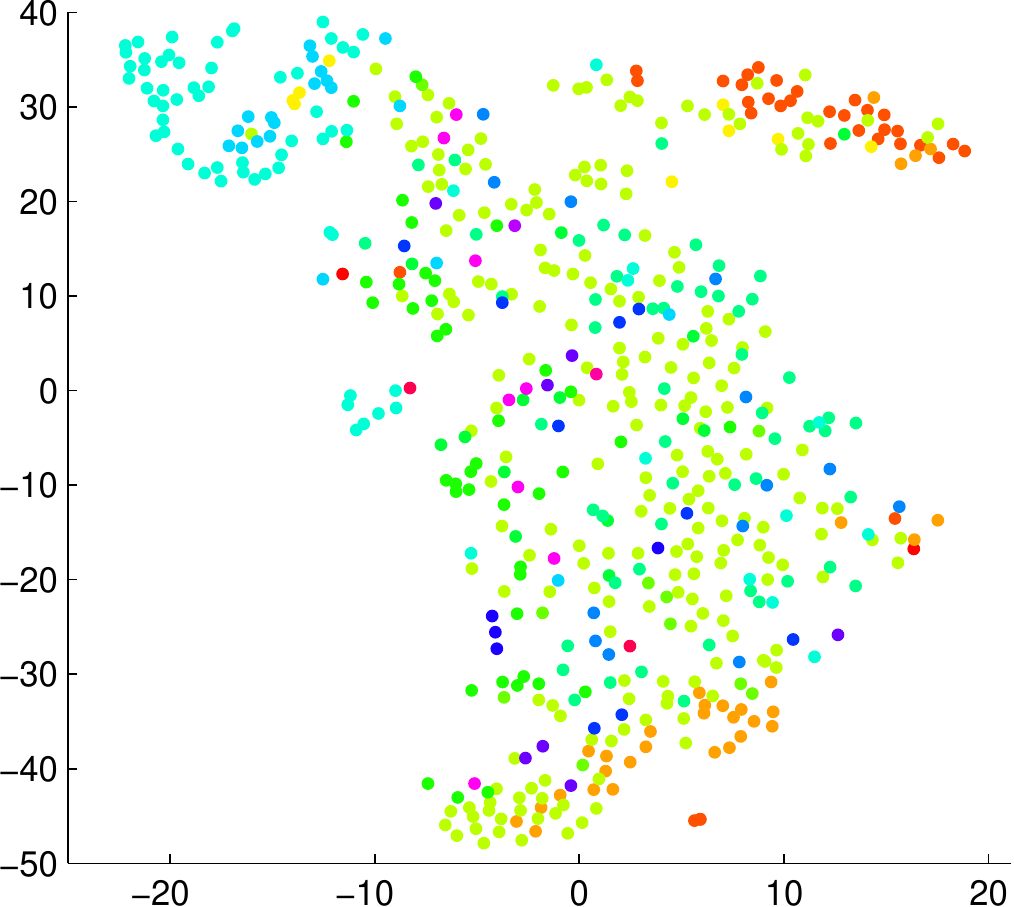}\label{subfig:U_tsne}
}
\hspace{0.5cm}
\subfigure[Adjacency matrix of rows of $U$ by correlation distance. Stock IDs are sorted by sectors.]{
\includegraphics[width=0.2\textwidth]{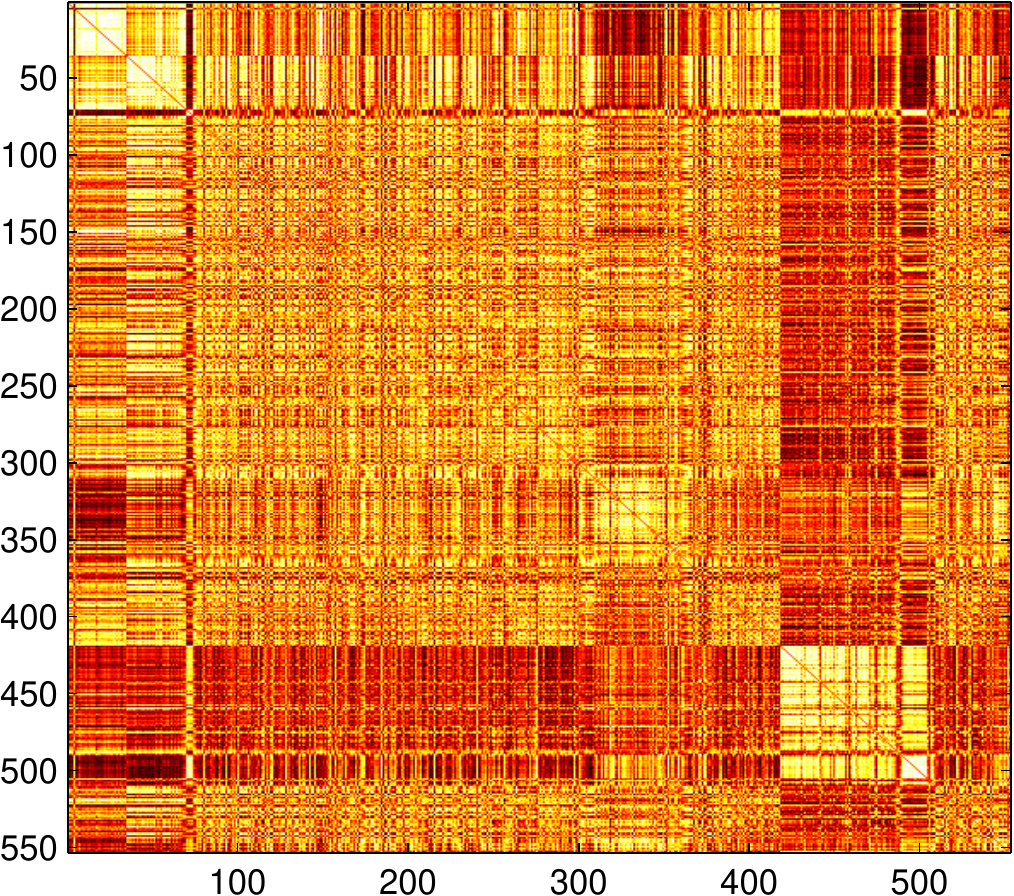}\label{subfig:U_adj}
}
\caption{\small Visualizing stocks.}
\label{fig:visualize_U}
\end{figure}

\myparab{Block structure of $U$.}
Given we have learnt $U$ with each row being the feature vector of a stock,
we study whether these vectors give meaningful interpretations by
applying t-SNE \cite{maaten08} to map our high-dimensional (10D) stock feature vectors on a low-dimensional (2D) space.
Intuitively, similar stocks should be close together in the 2D space, and by ``similar" we mean stocks being
in the same (or similar) sectors according to North American Industry Classification System (NAICS).
Figure \ref{subfig:U_tsne} confirms our supposition by having stocks of the same color, \ie in the same sector,
being close to each other. 
Another way to test $U$ is to compute the stock adjacency matrix.
Figure \ref{subfig:U_adj} shows the result with a noticeable block diagonal structure,
which independently confirms our claim that the learnt $U$ is meaningful.

Furthermore, we show the learnt $U$ also captures connections between stocks that are not captured by NAICS.
Table \ref{table:closest_stocks} shows the 10 closest stocks to Bank of America (BAC), Home Depot (HD) and Google (GOOG) according to $U$.
For BAC, all close stocks are in finance or insurance, \eg Citigroup (C) and Wells Fargo (WFC), and
can readily be deduced from NAICS.
However, the stocks closest to HD include both retailers, \eg Lowe's (LOW) and Target (TGT), and related non-retailers,
including Bemis Company (BMS, specializes in flexible packaging) and Vulcan Materials (VMC, specializes in construction materials).
Similarly, the case of GOOG reveals its connections to biotechnology stocks including Celgene Corporation (CELG) and
Alexion Pharmaceuticals (ALXN).
Similar results have also been reported by \cite{doyle09}.

\begin{table}
\small
\begin{center}
\caption{\small Closest stocks. Stocks are represented by ticker symbols.}
\label{table:closest_stocks}
\begin{tabular}{c | p{0.35\textwidth}}
\hline
Target & 10 closest stocks\\
\hline
BAC & XL STT KEY C WFC FII CME BK STI CMA \\
HD & BBBY LOW TJX BMS VMC ROST TGT AN NKE JCP \\
GOOG & CELG QCOM ORCL ALXN CHKP DTV CA FLIR ATVI ECL\\
\hline
\end{tabular}
\end{center}
\end{table}

\begin{figure}
\centering
\includegraphics[width=0.4\textwidth]{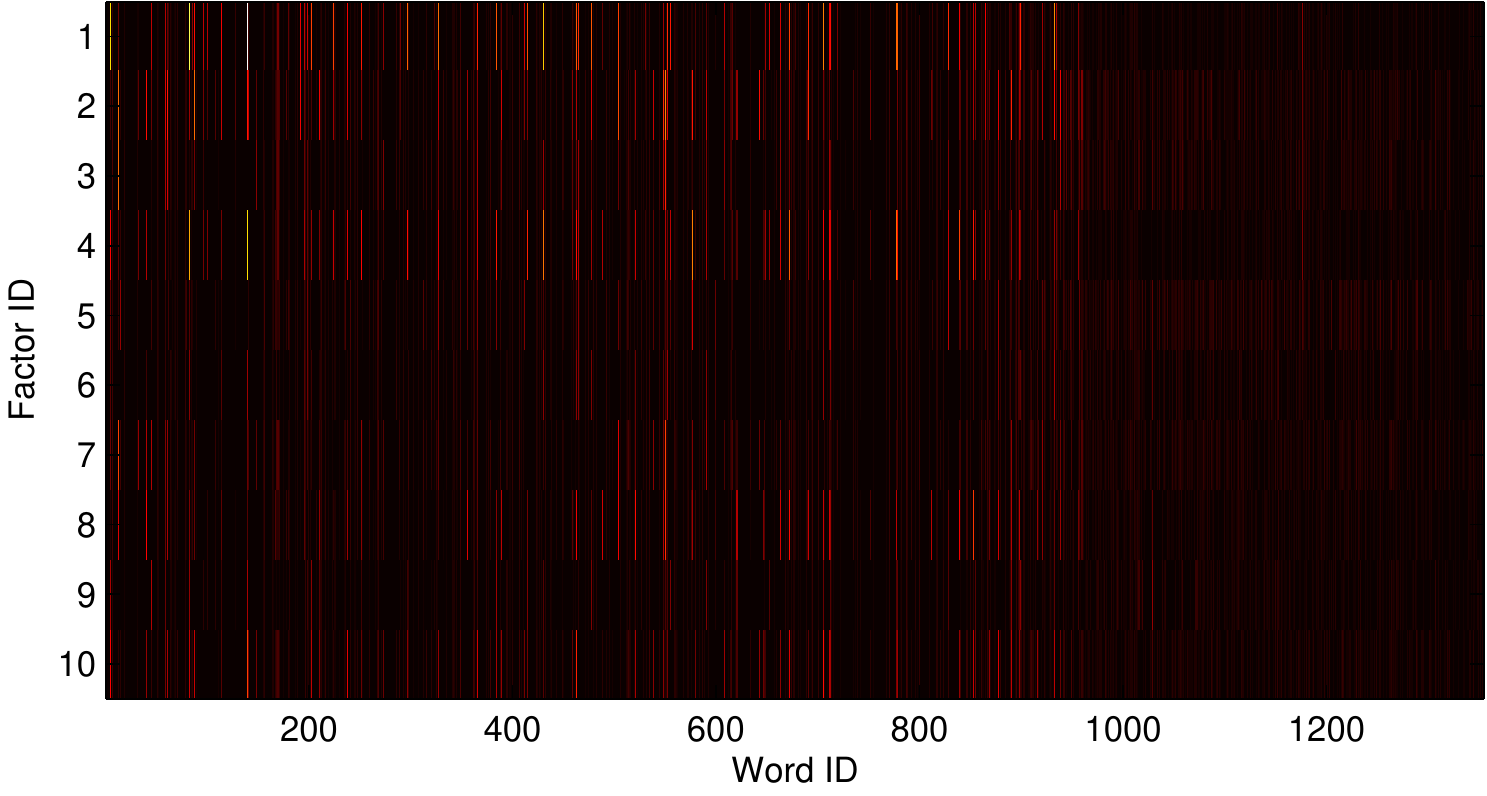}
\caption{\small Heatmap of $W$. It is inter and intra-column sparse.}
\label{fig:visualize_W}
\vspace{-0.6cm}
\end{figure}

\myparab{Sparsity of $W$.}
Figure \ref{fig:visualize_W} shows the heat map of our learnt $W$.
It shows that we are indeed able to learn the desired sparsity structure:
(a) few words are chosen (feature selection) as seen from few columns being bright, and
(b) each chosen word corresponds to few factors.

Studying $W$ reveals further insights on the stocks. We consider the ten most positive and negative words of two latent factors
as listed in Table \ref{table:posneg_words}. We note that the positive word list of one factor 
has significant overlap with the negative word list 
of the other factor. This leads us to hypothesize that the two factors are anticorrelated.

To test this hypothesis, we find the two sets of stocks that are dominant in one factor:\footnote{That is,
the stock's strength in that factor is in the top 40\% of all stocks \emph{and} its strength in the other factor is in the bottom 40\%.}
\{IRM, YHOO, RYAAY\} are dominant in factor 1, and \{HAL, FFIV, MOS\} are dominant in factor 2.
Then we pair up one stock from each set by the stock exchange from which they are traded: YHOO and FFIV from NASDAQ, and
IRM and HAL from NYSE. We compare the two stocks in a pair by their performance (in cumulative returns) relative to the
stock index that best summarizes the stocks in the exchange (\eg S\&P 500 for NYSE), 
so that a return below that of the reference index can be considered losing to the market,
and a return above the reference means beating the market.
Figure \ref{fig:opp_stocks} shows that two stocks with different dominant factors are in opposite beating/losing positions (relative to the reference index) for most of the time, and for the (IRM, HAL) pair the two stocks interchange beating/losing positions multiple times.

\begin{figure*}
\small
\centering
\subfigure[Stock weights from portfolio due to our strategy.]{\label{fig:visualize_B}
\includegraphics[width=0.4\textwidth]{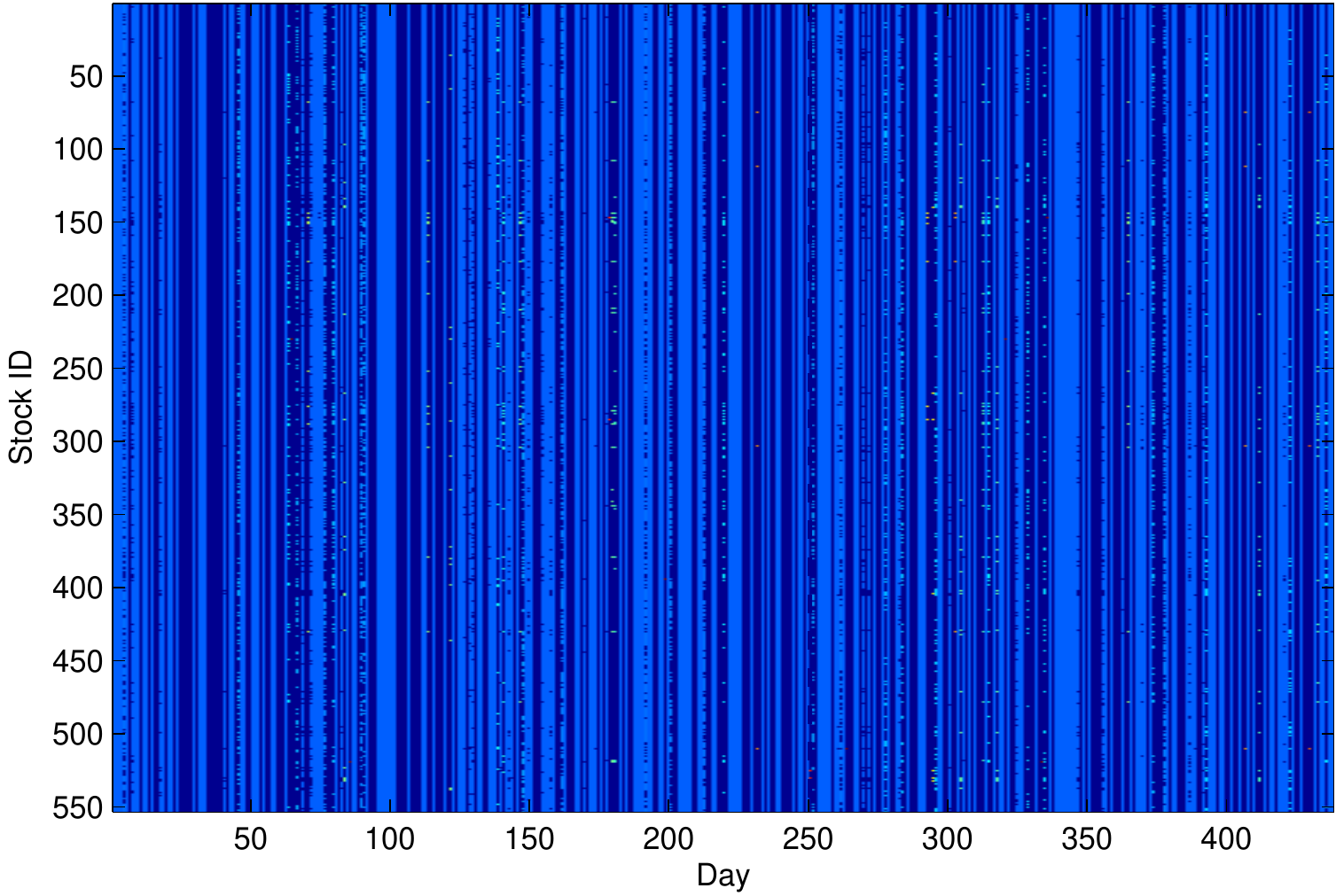}}
\subfigure[Cumulative returns.]{\label{fig:returns_1213}
\includegraphics[width=0.45\textwidth]{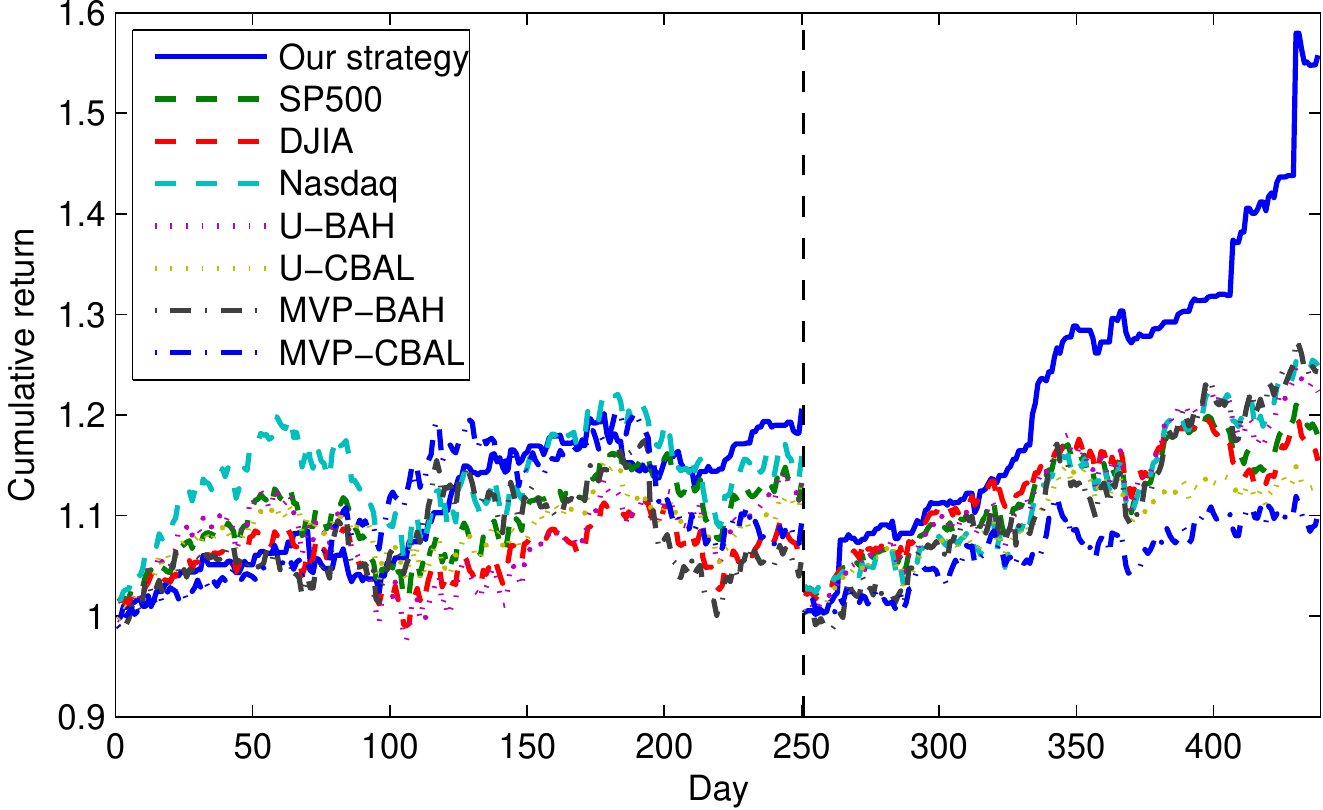}}
\caption{\small Visualizing our strategies and returns. Region left (right) of dashed line corresponds to 2012 (2013).}
\end{figure*}

\begin{table*}
\small
\begin{center}
\caption{\small Top ten positive and negative word lists of two factors.}
\label{table:posneg_words}
\begin{tabular}{c | c}
\hline
List & Words \\
\hline
Factor 1, positive & street billion goal designed corporate ceo agreement position buyers institute \\
Factor 1, negative & wall worlds minutes race free short programs university chairman opposition \\
Factor 2, positive & wall start opposition lines asset university built short race risks \\
Factor 2, negative & agreement designed billion tough bond set street goal find bush \\
\hline
\end{tabular}
\end{center}
\vspace{-0.3cm}
\end{table*}

\begin{figure}
\small
\centering
\includegraphics[width=0.4\textwidth]{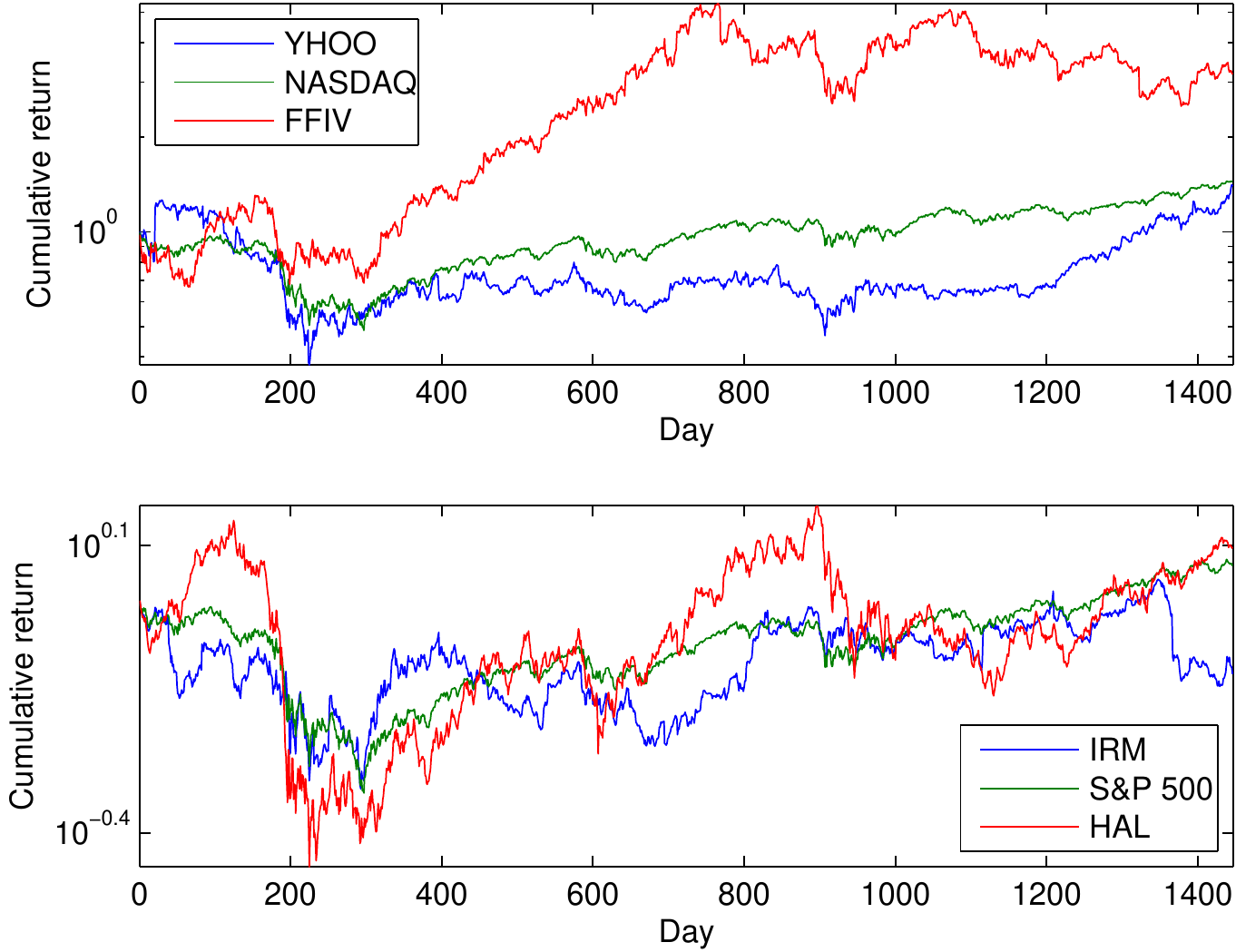}
\vspace{-0.2cm}
\caption{\small Returns of stocks with a different dominating factor. Green line is the reference index.}
\label{fig:opp_stocks}
\vspace{-0.5cm}
\end{figure}

\textbf{Visualizing learnt portfolio and returns.}
We try to gain a better understanding of our trading strategy by visualizing the learnt stock portfolio.
Figure \ref{fig:visualize_B} shows (bright means higher weight to the corresponding stock)
that our trading strategy alternates between three options on each day: (a) buy all stocks when an optimistic
market is expected, (b) buy no stocks when market pessimism is detected, and (c) buy a select set of stocks.
The numbers of days with (a) or (b) chosen are roughly the same, while that of (c) are fewer but still significant.
This shows our strategy is able to intelligently select stocks to buy/avoid in response to market conditions.

\myparab{Reaction to important market events.} To understand why our strategy results in better returns than the baselines, we also plot the cumulative returns
of the different trading strategies. Figure \ref{fig:returns_1213} reveals that our strategy is more stable
in growth in 2012, in that it avoids several sharp drops in value experienced by other strategies
(this can also be seen from the fact that our strategy has the lowest maximum drawdown and CVaR).
Although it initially performs worse than the other baselines (Nasdaq in particular), it is able to
catch up and eventually beat all other strategies in the second half of 2012.
It appears the ability to predict market drawdown is key for a good trading strategy using newspaper text (also see \cite{tetlock07}).

Looking deeper we find WSJ to contain cues of market drawdown for two of the five days in 2012 and 2013
that have S\&P 500 drop by more than 2\%.
On 6/1/2012, although a poor US employment report is cited as the main reason for the drawdown,
the looming European debt crisis may have also contributed to a negative investor sentiment,
as seen by ``euro" being used in many WSJ articles on that day. 
On 11/7/2012, the US presidential election results cast fears on a fiscal cliff and more stringent controls
on the finance and energy sectors. Many politics-related words, \eg democrats, election, %night,
won, voters, were prominent in WSJ on that day.

In 2013, our strategy is also able to identify and invest in rapidly rising stocks
on several days, which resulted in superior performance.
We note the performance of our algorithm in the two years are not the same, with 2013 being
a significantly better year.
To understand why, we look into the markets, and notice 2013 is an ``easier" year because
(a) other baseline algorithms also have better performance in 2013, and
(b) the volatility of stocks prices in 2012 is higher, which suggests the prices are ``harder" to predict.
In terms of S\&P 500 returns, 2012 ranks 10th out of 16 years since 1997, while 2013 is the best year
among them.

\section{Related Work}\label{sec:related}
Our discussion here focuses on works that study the connection between news texts (including those generated
from social media) and stock prices. Portfolio optimization (\eg \cite{markowitz52,cover91,borodin04,agarwal06,ganesh13} and references therein) is an important area in financial econometrics, but it is not directly relevant to our work because it does not incorporate news data.

The predictive power of news articles to the financial market has been extensively studied. 
\citet{tetlock07} applied sentiment analysis to a Wall Street Journal column and
showed negative sentiment signals precede a decline in DJIA.
\citet{chan03} studied newspaper headlines and showed investors tend to underreact to negative news.
\citet{dougal11} showed that the reporting style of a columnist is causally related to market performance.
\citet{wuthrich98,lavrenko00,fung02,schumaker09,zhang10} use news media to predict stock movement machine learning and/or data mining techniques. On top of using news, other text sources are also examined, such as corporate announcements \cite{hagenau12,mittermayer06},
online forums \cite{thomas00},
blogs \cite{zhang10},
and online social media \cite{zhang10,mao11}.
See \cite{minev12} for a comprehensive survey.

\myparab{Comparison to existing approaches.}
Roughly speaking, most prediction algorithms discussed above follow the same framework: first an algorithm constructs a feature vector based on the news articles. 
Next the algorithm will focus on prediciton on the subset of stocks or companies mentioned in the news. Different feature vectors are considered, \eg \cite{lavrenko00} use vanilla bag-of-word models while \cite{zhang10} extracts sentiment from text. Also, most ``off-the-shelf'' machine learning solutions, such as generalized linear models \cite{tetlock07}, Naive Bayes classifiers \cite{lavrenko00}, and Support Vector Machines \cite{schumaker09} are examined in the literature. Our approach differ from the existing ones in the following two ways:

\mypara{(1) No NLP.} Unlike \cite{tetlock07,schumaker09,hagenau12}, we do not attempt to interpret or understand news articles with techniques like sentiment analysis and named entity recognition. In this way, the architecture of our prediction algorithm becomes simpler (and thus has lower variance). 

\mypara{(2) Leveraging correlation between stocks.} \citet{lavrenko00,fung02} also make predictions without using NLP, but all these algorithms do not leverage the correlations that can exist between different stocks. It is not clear how these algorithms can be used to predict a large number of stocks without increasing model complexity substantially.

\section{Conclusion}

In this paper we revisit the problem of mining text data to predict the stock market.
We propose a unified latent factor model to model the joint correlation between
stock prices and newspaper content, which allows us to make predictions on individual
stocks, even those that do not appear in the news.
Then we formulate model learning as a sparse matrix factorization problem solved using ADMM.
Extensive backtesting using almost six years of WSJ and stock price data
shows our method performs substantially better than the market and a number of portfolio building strategies.
We note our methodology is generally applicable to all sources of text data,
and we plan to extend it higher frequency data sources such as Twitter.

{
\small
\bibliography{stock_news}
\bibliographystyle{IEEEtranN}
}

\end{document}